\documentclass{article} 
\PassOptionsToPackage{table}{xcolor}
\usepackage{iclr2027_conference,times}

\usepackage[T1]{fontenc}

\usepackage{amsmath,amsfonts,bm}

\def\eqref#1{equation~\ref{#1}}

\def\1{\bm{1}}

\DeclareMathAlphabet{\mathsfit}{\encodingdefault}{\sfdefault}{m}{sl}
\SetMathAlphabet{\mathsfit}{bold}{\encodingdefault}{\sfdefault}{bx}{n}

\usepackage{hyperref}
\usepackage{url}
\usepackage{xspace}
\usepackage{amssymb}
\usepackage{multirow}
\usepackage{graphicx}
\usepackage{booktabs}      
\usepackage[table]{xcolor}  
\definecolor{dynaweavebg}{RGB}{235,243,251} 

\title{\name: Learning to Adapt through\\ Dynamic Scenario Synthesis}

\author{%
  Xuan Yang\textsuperscript{1,2}\quad
  Chunxiao Liu\textsuperscript{3}\quad
  Hongsheng Xin\textsuperscript{4}\quad
  Hao Xu\textsuperscript{4}\quad
  Ning Miao\textsuperscript{1,2}%
}

\newcommand{\name}{DynaWeave\xspace}

\iclrfinalcopy 
\begin{document}

\maketitle
\fancyhead{}
\renewcommand{\headrulewidth}{0pt}

\begingroup
\renewcommand{\thefootnote}{}
\footnotetext{%
  \textsuperscript{1}Department of Data Science, City University of Hong Kong.\quad
  \textsuperscript{2}Hong Kong Institute of AI for Science, City University of Hong Kong.\quad
  \textsuperscript{3}Independent Researcher.\quad
  \textsuperscript{4}Li Auto Inc.%
}
\endgroup

\begin{abstract}
Large language model (LLM) agents increasingly use tools to complete tasks in stateful environments. However, many training environments remain largely static apart from changes caused by the agent’s own actions, whereas external events introduced by other users, services, or sudden contingencies in real environments can invalidate ongoing plans. To address this gap, we introduce \name{}, an automated synthesis framework that transforms existing static tasks into verifiable dynamic scenarios. The framework generates exogenous events relevant to each task, covering changes in task requirements, environment state, and external information. It uses directed acyclic graphs (DAGs) to represent dependencies between events and an execution engine to activate them during agent interaction. A dynamic verifier revises only the evaluation checks affected by events that actually occur and retains all other checks. Using \name{}, we collect 2,095 verified interaction trajectories across 36 environments and use them for supervised fine-tuning of Qwen3.5 models. Experiments on Gaia2, CostBench, and ToolMaze show that \name achieves consistent improvements in overall performance across all three evaluated model sizes, increasing the task completion rate on CostBench by 67.34 percentage points and even exceeding the performance of Gemini-3.1-Pro-Preview on ToolMaze. These results support dynamic scenario synthesis as a practical approach to training agents to adapt to changing environments. Our code is available at 
\url{https://github.com/Miaow-Lab/DynaWeave}.

\end{abstract}
\section{Introduction}
Large language models (LLM) are increasingly capable of using tools to complete tasks in external environments beyond the conversation. For example, an agent may search a database, compare available options, and then book a room or issue a refund through a tool. Existing benchmarks provide executable environments such as websites, applications, and tool sandboxes in which agents perform tasks, and automatically assess whether the results satisfy the user's request. ~\citep{zhou2024webarenarealisticwebenvironment,xie2024osworldbenchmarkingmultimodalagents,trivedi2024appworldcontrollableworldapps,lu2025toolsandboxstatefulconversationalinteractive,yao2024taubenchbenchmarktoolagentuserinteraction,barres2025tau2benchevaluatingconversationalagents}. Recent advances in environment synthesis further automate the construction of applications, tasks, and verifiers, expanding the supply of verifiable interactions for agent training and evaluation ~\citep{dong2026agentworldscalingrealworldenvironment,song2026envscalerscalingtoolinteractiveenvironments,wang2026agentworldmodelinfinity,froger2025arescalingagentenvironments}.

Despite this recent progress, most existing environments rely on an implicit assumption: once a task begins, the environment remains largely unchanged except through the actions of the evaluated agent. In practice, however, real-world environments offer no such guarantee of stability: another user may claim a resource that the agent had previously found available; a service may suddenly become unavailable; or an unexpected contingency may disrupt the original plan. We refer to such changes that affect a task during its execution but are not caused by the evaluated agent as \emph{exogenous events}. These changes can easily invalidate an otherwise correct plan, making success contingent on the agent's ability to dynamically revise its decisions rather than merely execute them.

Recent benchmarks have demonstrated the importance of evaluating agents under dynamic scenarios. Unlike previous static benchmarks, a dynamic scenario is characterized by the possibility of exogenous events during task execution. Gaia2~\citep{froger2026gaia2benchmarkingllmagents} provides asynchronous environments that evolve independently of the agent; CostBench~\citep{liu2026costbenchevaluatingmultiturncostoptimal} changes tool costs and availability during execution, requiring agents to revise their plans; ToolMaze~\citep{zhu2026toolsfailbenchmarkingdynamic} introduces explicit or implicit tool failures that require recovery and replanning; and ClawMark~\citep{meng2026clawmarklivingworldbenchmarkmultiturn} incorporates external updates into multi-turn, multi-day workflows. TThese benchmarks highlight the value of evaluating adaptation, but constructing dynamic scenarios and their corresponding verifiers can require substantial manual effort. Producing such scenarios at the scale needed for agent training therefore remains a challenge.


In this work, we propose \name, a framework that automatically transforms existing static tasks into verifiable dynamic scenarios. Given an executable environment, a task, and its original verifier, \name identifies possible changes relevant to the task and generates corresponding exogenous events. It filters out events that are invalid, implausible, or have no meaningful effect on the task. The resulting scenarios are configured before evaluation, and events are activated during agent interaction when their conditions are met. The original verifier is adapted to account for events that actually occur while preserving unaffected task requirements. This design combines automatic event generation, reproducible execution, and verification, allowing existing task collections such as EnvScaler~\citep{song2026envscalerscalingtoolinteractiveenvironments} and Agent-World~\citep{dong2026agentworldscalingrealworldenvironment} to be extended without manually designing dynamic scenarios for each domain.

To evaluate the value of these scenarios for agent training, we use \name to collect 2,095 verified interaction trajectories and subsequently perform supervised fine-tuning on Qwen3.5 models with 2B, 4B, and 9B parameters. Across all three model sizes, this training consistently improves the overall scores on Gaia2 and ToolMaze, as well as the task completion rate on CostBench. In particular, the Qwen3.5-2B model improves its task completion rate on CostBench from 3.00\% to 70.34\%, representing a substantial gain of 67.34 percentage points. Furthermore, on ToolMaze, the 9B model achieves a composite score of 32.28 after training, even exceeding the baseline performance of Gemini-3.1-Pro-Preview (32.19). These results strongly indicate that the benefits of training on the generated trajectories generalize well and extend beyond the training environments, effectively improving the models’ adaptability to unexpected changes during task execution.

The main contributions of our work are as follows: (1) We propose \name, a general framework for automatically transforming existing static, executable tool-use tasks into verifiable dynamic variants with exogenous changes occurring during task execution.(2) We construct 2,095 dynamic scenarios across 36 environments from EnvScaler, $\tau^{2}$-bench, ToolSandbox, and AgentDojo, and collect verified interaction trajectories for supervised fine-tuning. (3) Experiments on three dynamic benchmarks show consistent improvements across three model sizes, demonstrating the value of \name for training agents to adapt to changes in dynamic environments.

\section{Related Work}
\subsection{Executable benchmarks and environment synthesis}
Executable benchmarks evaluate agents through continuous interactions with working applications and rigorous checks of task completion. WebArena~\citep{zhou2024webarenarealisticwebenvironment} and OSWorld~\citep{xie2024osworldbenchmarkingmultimodalagents} support reproducible interaction with complex web and computer systems; AppWorld~\citep{trivedi2024appworldcontrollableworldapps} and ToolSandbox~\citep{lu2025toolsandboxstatefulconversationalinteractive} evaluate intermediate environment states or complete interaction trajectories; and $\tau$-bench~\citep{yao2024taubenchbenchmarktoolagentuserinteraction} incorporates simulated users and specific domain policies. AndroidWorld~\citep{rawles2025androidworlddynamicbenchmarkingenvironment} expands overall task coverage through parameterized initialization, which effectively varies task instances rather than necessarily introducing unpredictable external changes during execution. $\tau^2$-bench~\citep{barres2025tau2benchevaluatingconversationalagents} permits both the user and the agent to modify a shared environment, emphasizing real-time coordination between participants. Ultimately, these existing settings provide valuable, reusable tools and evaluation procedures on which more dynamic extensions can readily build.

Environment synthesis addresses the cost of constructing such resources. AgentScaler~\citep{fang2025generalagenticintelligenceenvironment} and AutoForge~\citep{cai2025autoforgeautomatedenvironmentsynthesis} automate environment construction for agent training. EnvScaler~\citep{song2026envscalerscalingtoolinteractiveenvironments} separates executable environment skeletons from task scenarios and rule-based validators, while Agent World Model (AWM)~\citep{wang2026agentworldmodelinfinity} constructs database-backed environments with accessible state for reward computation. Agent-World~\citep{dong2026agentworldscalingrealworldenvironment} further connects environment and task synthesis to capability gaps identified during training. Their primary focus is expanding the supply and diversity of executable tasks. \name addresses a complementary dimension: generating changes within an existing task, determining when and how they become observable, and adapting evaluation to their effects. It reuses the source environment and tool interface, preserving unaffected requirements rather than constructing every dynamic scenario from scratch. This executable foundation also allows event application to be checked separately from agent behavior, so errors in the generated scenario need not be treated as failures of adaptation.

\subsection{Dynamic and asynchronous agent benchmarks}
Dynamic benchmarks examine how agents respond when earlier information or plans become invalid. Gaia2~\citep{froger2026gaia2benchmarkingllmagents} introduces independently evolving, asynchronous environments, supported by ARE's~\citep{froger2025arescalingagentenvironments} for applications, event execution, and verification. ClawMark~\citep{meng2026clawmarklivingworldbenchmarkmultiturn} incorporates external updates into workflows spanning multiple turns and days. CostBench~\citep{liu2026costbenchevaluatingmultiturncostoptimal} studies economical replanning under changes such as tool failures and revised costs, whereas ToolMaze~\citep{zhu2026toolsfailbenchmarkingdynamic} combines tool dependency structures with explicit or implicit failures that may be transient or permanent. ClawArena~\citep{ji2026clawarenabenchmarkingaiagents} focuses on revising beliefs as evidence changes, and SimuHome~\citep{seo2026simuhometemporalenvironmentawarebenchmark} models temporal changes in a smart-home simulator. Together, these benchmarks expose distinct challenges in adaptation, from recognizing unreliable information to recovering from disrupted execution.

Our focus is complementary to constructing these evaluation settings: automatically deriving plausible events and corresponding verification rules from existing static tasks across domains. An execution engine enables scheduled changes, but event selection and evaluation must still account for the source task's requirements. ProEvolve~\citep{li2026worldwontstaystill} and EvoArena~\citep{xu2026evoarenatrackingmemoryevolution} address change across environment versions and successive tasks, including evolving tools, schemas, data, and agent memory. In contrast, \name retains the tool interface and environment rules while introducing changes during task execution. This scope supports paired evaluation of static and dynamic variants and the collection of demonstrations of adaptation within the same underlying environments.
\section{\name}

\begin{figure}[t]
    \centering
    \includegraphics[width=\linewidth]{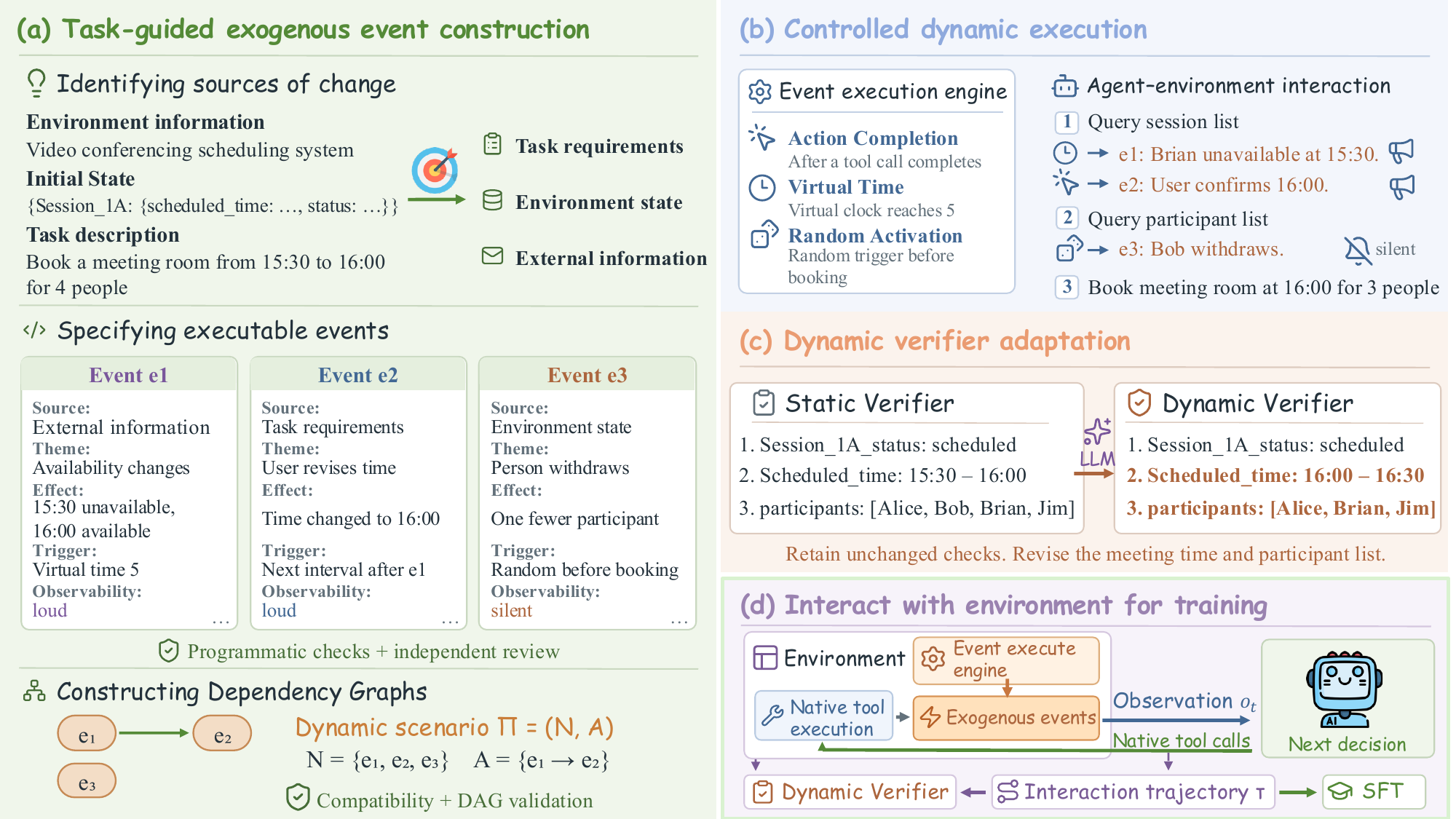}
    \caption{Overview of \name. (a) The framework generates task-grounded exogenous events from an existing static task, validates their specifications, and organizes compatible events into a dependency graph. (b) An execution engine activates events during agent interaction through action completion, virtual time, or random activation. (c) The dynamic verifier adapts checks affected by realized events while retaining unaffected requirements. (d) Agents interact through the original tools, receiving tool results and permitted event notifications; verified teacher trajectories provide demonstrations for supervised fine-tuning.}
    \label{fig:dynaweave_overview}
\end{figure}

\name{} transforms existing static tasks into verifiable dynamic scenarios by reusing their executable environments and adapting the original success criteria. Given an environment, its initial state, a task instruction, and the original verifier, it constructs a dynamic scenario $\Pi$ and a corresponding dynamic verifier $V_{\Pi}$. As illustrated in Figure~\ref{fig:dynaweave_overview}, the framework consists of three components. First, \emph{exogenous event construction} identifies plausible event themes relevant to the task, specifies each event's effects, activation conditions, and observability, and organizes compatible events into a directed acyclic graph (DAG). Second, \emph{controlled dynamic execution} activates events based on action completion, virtual time, or random triggers, while respecting event dependencies and environment conditions. Third, \emph{dynamic verifier adaptation} revises checks affected by events that actually occur and preserves requirements that remain applicable. When correctness depends on the timing or order of actions, the verifier also checks the execution trace.

\subsection{Exogenous Event Construction}
To construct plausible exogenous events for static tasks, we design a three-stage pipeline that first identifies event themes, then generates their specific content, and finally organizes the resulting events into a directed acyclic graph.

\textbf{Identifying Sources of Change.} 
The first stage identifies where relevant dynamics may arise during task execution. To guide the automatic generation of plausible event themes, we distinguish three sources of dynamics: (i) task requirements, where the user revises the original goal; (ii) environment state, where resource availability or tool functionality changes; and (iii) external information, where a contingency affect the goal or path of the task. For example, a typhoon warning at the intended destination may require the agent to reconsider a flight booking. Given the environment information, initial state, and task description, \name determines which sources can plausibly affect the task and proposes corresponding event themes. Each theme provides a concise description of a potential change, such as the user cancelling a trip, fewer rooms being available, or a typhoon warning being issued. These themes guide the generation of concrete events in the next stage.

\textbf{Specifying Executable Events.} 
Building on the selected themes, \name expands each concise description into a concrete event with three components. First, it specifies the source of the change, the environment entities or resources affected, and the resulting effects. Second, it defines the conditions under which the event is triggered, such as after a particular agent action or when a virtual clock reaches a scheduled time. Detailed trigger mechanisms and their implementation are discussed in Section~\ref{sec:Controlled Dynamic Execution}. Third, it specifies how the event becomes observable to the agent. We distinguish three modes: \emph{loud} events explicitly notify the agent of the change; \emph{silent} events provide no notification and must be discovered through available tools; and \emph{partial} observations reveal only part of the change, requiring further investigation to establish its implications. Generated event candidates undergo programmatic checks and independent review to assess whether their effects are executable, their causes are plausible, and their consequences are relevant to the task. Accepted events are then passed to the composition stage.

\textbf{Constructing Dependency Graphs.} 
In the final stage, \name{} combines multiple events into a coherent dynamic scenario. Some events depend on earlier changes; for example, a service can recover only after it has failed. An LLM selects compatible events and proposes dependencies based on their causes, effects, and activation conditions. The resulting scenario is represented as a directed acyclic graph (DAG), $\Pi = (\mathcal{N}, \mathcal{A})$, where each node in $\mathcal{N}$ represents an event. An edge $(e_i, e_j) \in \mathcal{A}$ means that event $e_i$ must occur before event $e_j$ can be triggered. The graph does not impose a fixed order on independent events. Its acyclic structure prevents circular dependencies in which events wait for one another. Each proposed graph is validated for structural correctness and consistency among event dependencies, activation conditions, and effects.

\subsection{Controlled Dynamic Execution}
\label{sec:Controlled Dynamic Execution}

Once the event DAG is constructed, \name uses a programmatic execution engine to monitor agent interactions and apply events when their trigger conditions are satisfied. We consider three trigger mechanisms: completion of a specified agent operation, passage of virtual time, and random activation to model the complex mechanisms of changes in real environment.

\textbf{Action Completion.} An event can be triggered after the agent completes a specified tool operation. The execution engine matches the completed call against the operation recorded in the event specification, including relevant arguments or occurrence counts when needed. This places changes at meaningful points in the task, such as immediately after the agent checks a resource's availability. The event is applied after the tool call finishes, so the operation itself remains intact while its result may become outdated before the agent's next decision. For example, after the agent confirms that a hotel room is available, an external booking may reserve that room, requiring the agent to reconsider its choice before completing the reservation.

\textbf{Virtual Time.} A virtual clock provides a strictly reproducible measure of elapsed time within the simulated environment. Starting from an initial state of $c_0=0$, the clock advances after each executed tool operation according to $c_{k+1}=c_k+d(a_k)$, where $d(a_k)$ is the simulated duration of action $a_k$. We use the durations pre-defined by the source environment when available; otherwise, we assign fixed durations to tool operations and keep them strictly consistent across all agent evaluations. To precisely determine an event's scheduled time $c_e$, we accumulate these simulated action durations along the task's reference static trajectory and select a strategic time when the event's prerequisite conditions hold and its effects can still influence subsequent decisions. During dynamic execution, the event is automatically activated at the first interaction boundary where $c_k \geq c_e$ and its eligibility conditions are satisfied.

\textbf{Random Activation.} At each interaction interval, an event that satisfies its eligibility conditions can be activated with a specified probability $p_e$. We fix the random seed before execution to make the random activation process reproducible. An event may remain inactive when the interaction ends, and the verifier accounts only for changes that actually occurred.

Under all three mechanisms, each event is applied at most once. When an event occurs, the engine applies its specified effects, provides only the observations allowed by its observability mode, and records the changes separately from those caused by the agent's tool calls. These records support the dynamic verifier described in Section~\ref{sec: Dynamic Verifier Adaptation}.

\subsection{Dynamic Verifier Adaptation}
\label{sec: Dynamic Verifier Adaptation}
Given the original verifier $V_0$ and the generated event specifications, \name uses an LLM to identify checks whose validity depends on the information, state, or task requirements affected by each event. Unaffected checks are retained, while those whose expected outcomes have changed are revised accordingly. A task requirement is removed only when the user explicitly cancels it. Changes that affect how a task can be completed do not, by themselves, alter the requested outcome. For example, consider a meeting for which the user changes only the requested start time from 15:30 to 16:00. The original verifier would reject a booking at the updated time, whereas removing the time check could accept bookings at an incorrect time. The dynamic verifier instead revises the expected time while retaining other requirements. By contrast, a room becoming unavailable does not, by itself, authorize a change in meeting time. This distinction separates changes in the task from changes in the plan used to complete it.

The proposed modifications are expressed as structured rules and validated before execution for consistency with the task requirements and event specifications. Incompatible modifications are rejected at this stage. During execution, each modification takes effect only when its associated event occurs; inactive events leave the applicable checks unchanged. When multiple events occur, their modifications are applied in the realized execution order, consistent with the DAG. The verifier therefore evaluates the requirements applicable to the observed execution rather than assuming that all potential changes have occurred. When a requirement depends on the timing or order of actions, \name supplements outcome checks with checks over the execution trace. These checks assess whether the agent's actions satisfy the requirements in effect at the relevant point in execution, while allowing different valid action sequences rather than prescribing a fixed solution.

Before accepting a generated verifier, \name{} checks that it agrees with the original verifier when events are disabled. Strict validation additionally tests two solutions: a solution to the original static task that the event is expected to invalidate, and an adapted solution that satisfies the revised requirements. The verifier should reject the first and accept the second. This comparison provides evidence that the event meaningfully affects the task and that the adapted verifier distinguishes an invalid response from a valid adaptation.

\subsection{Interaction with Dynamic Scenarios}
\label{sec:trajectory_collection}
\name{} retains the tools, argument schemas, and interaction protocol of the source benchmark. Dynamic scenarios change how the environment evolves and what the agent observes, without requiring a new interaction interface.

We model the interaction as a partially observable Markov decision process (POMDP). The action space remains unchanged, while the hidden state is extended to $x_t = (S_t, g_t, z_t)$. Here, $S_t$ contains the environment state and external information, $g_t$ denotes the current task requirements, and $z_t$ records the virtual clock, the events that have occurred, and the action counts used to determine when subsequent events are triggered.

At decision step $t$, let $h_t = (u, o_0, a_0, \ldots, o_{t-1}, a_{t-1})$ denote the history consisting of the initial task instruction $u$ and all preceding observations and actions. The current observation $o_t$ contains newly received tool results, user messages, and any event notifications permitted by the specified observability modes. The agent selects its next action based on both $h_t$ and $o_t$:
\begin{equation}
a_t \sim \pi_{\theta}(\cdot \mid h_t, o_t),
\qquad
(x_{t+1}, o_{t+1}) \sim K_{\Pi}(\cdot \mid x_t, a_t),
\end{equation}
where $\pi_{\theta}$ is the agent's policy and $K_{\Pi}$ jointly specifies the next state and observation, accounting for tool execution and exogenous events. After a tool call, an event may change $S_t$ or, when authorized, revise $g_t$ before the agent's next decision. The agent must therefore use subsequent observations to determine whether its earlier information and plan remain valid. At the end of the interaction, $V_{\Pi}$ evaluates whether the agent has satisfied the task requirements, taking into account the events that actually occurred.

For supervised fine-tuning, we collect trajectories from a teacher model. These trajectories record the observations available to the teacher, including event notifications, user messages, or changes in tool results, together with its subsequent actions. They provide demonstrations of how the teacher responds to changes and adjusts its actions during task execution.
\section{Experiments}

\begin{figure}[t]
    \centering
    \includegraphics[width=\linewidth]{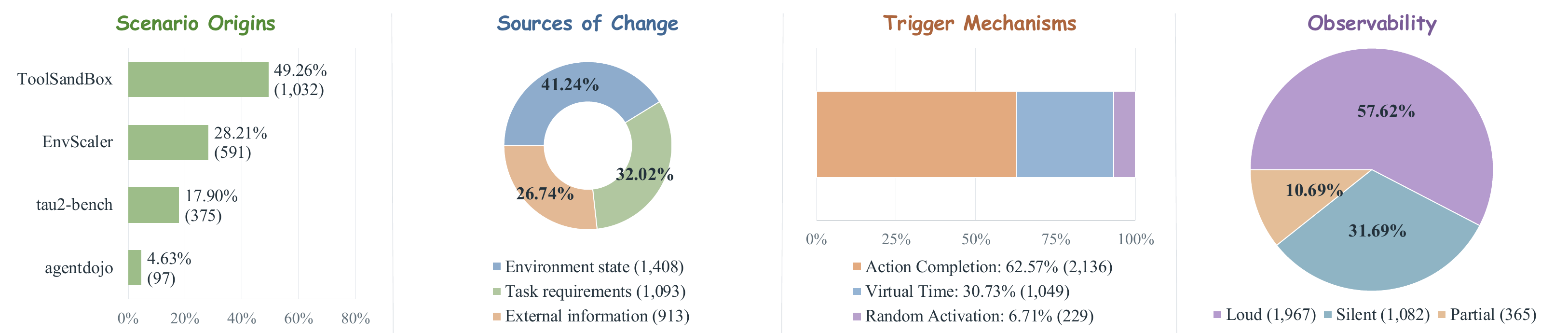}
    \caption{Statistics of the dynamic scenario corpus generated by \name.}
    \label{fig:statistic}
\end{figure}

\subsection{Experiment Setup}
\label{sec:experiment_setup}
\textbf{Training Details.} 
We apply \name to EnvScaler~\citep{song2026envscalerscalingtoolinteractiveenvironments} ToolSandbox~\citep{lu2025toolsandboxstatefulconversationalinteractive}, AgentDojo~\citep{debenedetti2024agentdojodynamicenvironmentevaluate}, and $\tau^2$-bench~\citep{barres2025tau2benchevaluatingconversationalagents}, extending their executable environments with dynamic scenarios. The resulting corpus contains 2,095 dynamic scenarios and 3,414 events across 36 environments, covering healthcare, scheduling, mobile assistance, workplace applications, travel, retail, banking, and telecommunications. Figure~\ref{fig:statistic} summarizes the corpus statistics. Using Qwen3.5-35B-A3B as the teacher model, we collect 2,095 verified interaction trajectories for supervised fine-tuning (SFT) of Qwen3.5 models. We fine-tune the models in non-thinking mode using LLaMA-Factory and LoRA with a rank of 16. All models are trained for two epochs with a learning rate of $1 \times 10^{-4}$, an effective batch size of 64, and a maximum sequence length of 16,384 tokens.

\textbf{Evaluation Benchmarks.} 
We evaluate \name on three dynamic benchmarks to assess generalization beyond the training environments. Gaia2~\citep{froger2026gaia2benchmarkingllmagents} provides asynchronous application environments that evolve independently of the agent. It evaluates execution, search, ambiguity resolution, adaptation, and reasoning under time constraints. CostBench~\citep{liu2026costbenchevaluatingmultiturncostoptimal} evaluates planning in a travel environment where tools have different costs. Its dynamic conditions change tool costs, user preferences, execution lengths, or tool availability, requiring agents to revise their plans while controlling cost. ToolMaze~\citep{zhu2026toolsfailbenchmarkingdynamic} evaluates recovery from explicit tool errors and from corrupted outputs without explicit error messages. Both types of failure can be transient or permanent.

\subsection{Main Results}
\label{sec:main_results}
Table~\ref{tab:main_results} presents the performance of \name on three dynamic benchmarks. Across all three model sizes, \name improves overall scores on Gaia2 and ToolMaze, as well as task completion on CostBench. These gains encompass more reliable tool use, more efficient task completion, and better aggregate recovery from tool failures, demonstrating that \name enhances agents' ability to adapt to dynamic scenarios and generalize beyond the training environments.

\textbf{\name transfers across distinct forms of environmental change.}
The three benchmarks test different dynamic scenarios, from asynchronous application updates to changing planning constraints and tool failures. \name yields gains across these settings: for 4B, Gaia2 overall improves from 3.3 to 5.0, CostBench completion rises from 72.41\% to 90.81\%, and ToolMaze success under perturbations increases from 21.25\% to 25.38\%. Both the 2B and 9B models likewise improve across all three benchmarks. \name-9B even surpasses GPT-4o and Gemini-3.1-Pro-Preview on ToolMaze. These consistent improvements across out-of-domain environments show that \name learn to adjust the actions and plans in response to dynamic changes, rather than merely reproduce tool-use patterns from the training environments.

\textbf{The gains extend to both execution reliability and efficiency.}
\name improves task completion without sacrificing execution efficiency. On CostBench, the 9B model raises completion to 92.39\% while reducing its cost gap from 0.755 to 0.140 and invalid tool use from 7.08\% to zero. ToolMaze likewise shows higher mean recovery rates and lower normalized recovery costs at every model size (The detailed results of ToolMaze are deferred to Appendix~\ref{app:toolmaze_detailed_results}), alongside improvements in Gaia2 Search and Execution. These joint gains are consistent with agents obtaining updated information when needed and revising their actions, rather than continuing a plan based on outdated assumptions. A targeted query after a change can incur an immediate cost but prevent repeated failed calls or actions that no longer advance the task. Responding appropriately to the new information can therefore improve both the likelihood of completion and the efficiency of reaching it.

\textbf{\name improves models across different sizes.}
\name consistently improves models of all evaluated sizes. Larger models retain stronger overall performance after training across the three benchmarks, suggesting that their underlying reasoning and tool-use capabilities may help them interpret changes and select appropriate responses. Targeted training also narrows the gap between model sizes: the trained 4B model outperforms the original 9B model on both Gaia2 overall and CostBench task completion. These results show that \name can strengthen smaller models while providing further gains for larger ones.

Despite these gains, models trained with \name still struggle when tool results are incorrect but contain no explicit error message, or when success requires responding to changes at the right time, as reflected in ToolMaze's corrupted-output conditions and Gaia2's Adaptability and Time categories. These cases require agents to recognize when information is unreliable or outdated and decide when to check it again. The limitations may stem from the absence of these types of changes in the dynamic scenarios used for training.

\begin{table}[t]
\centering
\setlength{\belowcaptionskip}{\baselineskip}
\small
\setlength{\tabcolsep}{3pt}
\caption{Main results on Gaia2, CostBench, and ToolMaze. For CostBench, Cost Gap measures excess tool execution cost relative to the optimal reference; TCR (Task Completion Ratio) is the percentage of correctly completed tasks; and ITUR (Invalid Tool-Use Ratio) is the percentage of invalid tool calls. ToolMaze Avg denotes its official composite score. Detailed ToolMaze results and the definition of Avg are provided in Appendix~\ref{app:toolmaze_detailed_results}.}
\label{tab:main_results}
\resizebox{\linewidth}{!}{%
\begin{tabular}{lrrrrrrrrrr}
\toprule
 & \multicolumn{6}{c}{Gaia2} & \multicolumn{3}{c}{CostBench} & ToolMaze \\
\cmidrule(lr){2-7}\cmidrule(lr){8-10}\cmidrule(lr){11-11}
Model & Execution & Search & Ambiguity & Adaptability & Time & Overall & Cost Gap $\downarrow$ & TCR & ITUR $\downarrow$ & Avg \\
\midrule
\multicolumn{11}{l}{\textit{Advanced models}} \\
GPT-5.5 & 79.4 & 96.2 & 46.9 & 55.0 & 4.4 & 56.4 & 0.000 & 95.01 & 0.00 & 35.94 \\
Gemini-3.1-Pro-Preview & 78.6 & 92.8 & 40.6 & 45.9 & 2.1 & 52.0 & 0.010 & 95.28 & 1.42 & 32.19 \\
GLM-5.1 & 71.2 & 83.8 & 39.4 & 56.9 & 1.2 & 50.5 & 0.030 & 93.18 & 0.29 & 32.28 \\
GPT-4o & 8.3 & 17.5 & 4.4 & 6.2 & 5.8 & 7.4 & 1.179 & 90.29 & 19.96 & 31.29 \\
\midrule
Qwen3.5-2B & 0.0 & 2.1 & 0.0 & 0.0 & 0.0 & 0.4 & 1.998 & 3.00 & 19.81 & 18.52 \\
\rowcolor{dynaweavebg}
\quad +\name & \textbf{1.0} & \textbf{5.2} & 0.0 & 0.0 & 0.0 & \textbf{1.3} & \textbf{1.242} & \textbf{70.34} & \textbf{0.16} & \textbf{23.49} \\
\addlinespace[2pt]
Qwen3.5-4B & 6.3 & 10.4 & 0.0 & 0.0 & 0.0 & 3.3 & 0.650 & 72.41 & 13.98 & 26.26 \\
\rowcolor{dynaweavebg}
\quad +\name & \textbf{8.3} & \textbf{14.6} & \textbf{2.1} & 0.0 & 0.0 & \textbf{5.0} & \textbf{0.331} & \textbf{90.81} & \textbf{2.47} & \textbf{29.99} \\
\addlinespace[2pt]
Qwen3.5-9B & 6.3 & 16.7 & 0.0 & 0.0 & 0.0 & 4.6 & 0.755 & 86.76 & 7.08 & 30.59 \\
\rowcolor{dynaweavebg}
\quad +\name & \textbf{10.4} & \textbf{26.0} & \textbf{3.1} & \textbf{2.1} & \textbf{1.0} & \textbf{8.5} & \textbf{0.140} & \textbf{92.39} & \textbf{0.00} & \textbf{32.28} \\
\bottomrule
\end{tabular}%
}
\end{table}

\subsection{Data Efficiency and Performance Saturation}
\label{sec:data_efficiency}

We examine how the number of training trajectories affects Qwen3.5-4B performance, keeping the source tasks, environments, and number of training epochs fixed. Figure~3 shows that adding trajectories improves task completion on CostBench and both task success under perturbations and the composite score on ToolMaze. A substantial share of these gains is achieved with 1,048 trajectories, approximately half of the full corpus. At this size, CostBench task completion reaches 87.08\% and the ToolMaze composite score reaches 28.91, compared with 90.81\% and 29.99, respectively, when all 2,095 trajectories are used. The gains become smaller as the training set grows.

The effects on other metrics are less uniform. ToolMaze's normalized recovery cost decreases across all evaluated training set sizes. However, CostBench's invalid tool use ratio and cost gap reach their lowest values before the full corpus is used, while ToolMaze's PRR peaks at 1,571 trajectories and then declines slightly. These results show that higher task completion does not necessarily coincide with improvements in every measure of reliability and efficiency. One possible explanation is that SFT trains the model to imitate demonstrated actions without directly optimizing execution cost. Overall, training on more trajectories generated with \name{} improves task performance within the evaluated range, but the gains diminish and additional data does not improve all metrics consistently.

\begin{figure}[t]
\centering
\includegraphics[width=\linewidth]{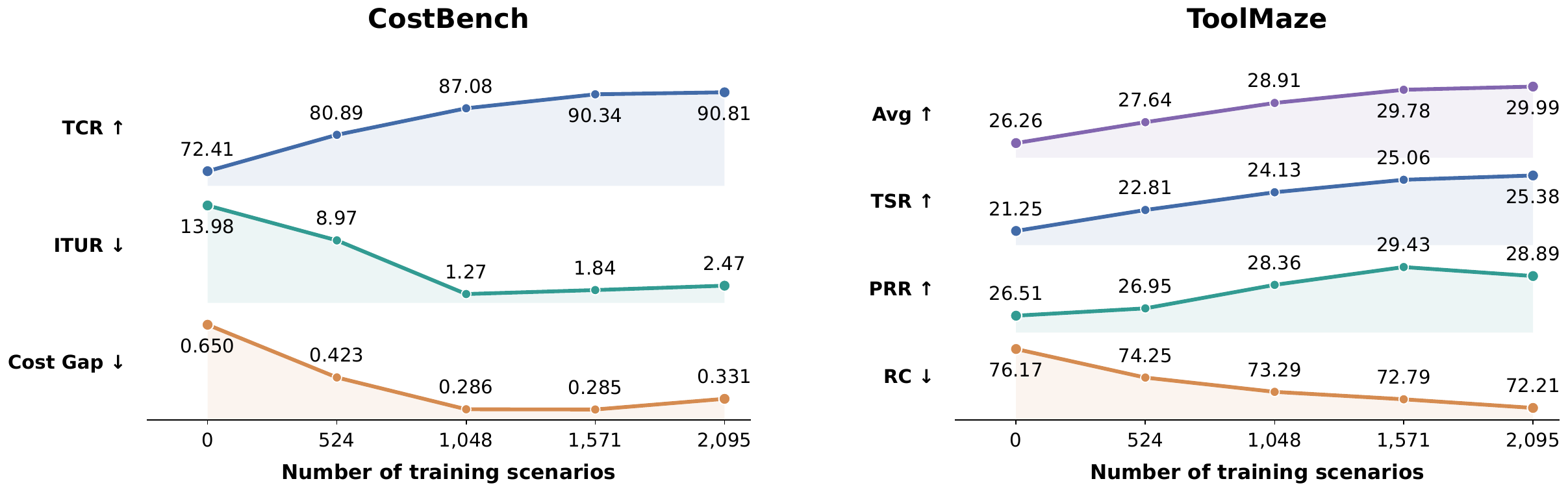}
\caption{Effect of training scenario quantity on Qwen3.5-4B performance on CostBench (left) and ToolMaze (right). Zero scenarios denotes the baseline Qwen3.5-4B.}
\label{fig:data_efficiency}
\end{figure}

\subsection{Contributions of Different Event Sources}
\label{sec:event_ablation}
To examine the contributions of different sources of change, we compare training on a mixture of all three sources with training on subsets that each exclude one source: \emph{Task Requirement}, \emph{Environment State}, or \emph{External Information}. For this comparison, we use Qwen3.5-4B and 1,048 trajectories for each setting. As shown in Table~2, the mixed subset achieves the highest task completion rate on CostBench and the highest composite score on ToolMaze. It also yields the highest ToolMaze task success under perturbations and the lowest recovery cost. CostBench task completion drops most when Task Requirement events are excluded, whereas ToolMaze success under perturbations and recovery efficiency decline most when Environment State events are excluded. These patterns suggest complementary benefits from learning to revise task goals and adjust execution to changes in the environment. In particular, demonstrations involving changes in task progress may help agents identify which actions remain necessary and avoid redundant execution during recovery.

Compared with the mixed subset, excluding External Information events improves ToolMaze task success without injected failures, but lowers average success and increases recovery cost under tool failures or corrupted outputs. This pattern suggests that learning to respond to new information may help agents adjust their actions when execution is disrupted. On CostBench, some subsets that omit an event source achieve a lower cost gap or a lower invalid tool use ratio than the mixed subset, but all have lower task completion rates. These results highlight the need to assess efficiency alongside task completion.

We further compare these subsets with a smaller mixed subset containing 524 trajectories, approximately 25\% of the full training corpus. On ToolMaze, this smaller subset achieves a composite score of 27.64, exceeding all three subsets that contain 1,048 trajectories but omit one source. It also achieves higher average task success under perturbations and lower recovery cost than these larger subsets, despite lower success without injected failures. This comparison highlights the importance of data composition as well as quantity. However, differences in task and environment coverage, together with the different numbers of optimization steps across dataset sizes, prevent us from attributing this advantage solely to the inclusion of all three event sources. Overall, the results support retaining all three sources to improve task completion and recovery, rather than increasing data quantity without considering the diversity of changes it covers.

\begin{table}[t]
\centering
\setlength{\belowcaptionskip}{\baselineskip}
\small
\setlength{\tabcolsep}{3pt}
\caption{Effect of event-source composition on Qwen3.5-4B performance. Data denotes the number of training trajectories; Mixed includes all three event sources. For ToolMaze, NP (No Perturbation) reports task success without injected failures; TSR (Task Success Rate) reports average task success under tool failures or corrupted outputs; and RC (Recovery Cost) reports average normalized recovery cost under the same conditions. Bold marks the best value in each metric column.}
\label{tab:event_ablation}
\begin{tabular}{lrrrrrrrr}
\toprule
 & & \multicolumn{3}{c}{CostBench} & \multicolumn{4}{c}{ToolMaze} \\
\cmidrule(lr){3-5}\cmidrule(lr){6-9}
Training subset & Data & Gap $\downarrow$ & TCR $\uparrow$ & ITUR $\downarrow$ & NP $\uparrow$ & TSR $\uparrow$ & RC $\downarrow$ & Avg $\uparrow$ \\
\midrule
Mixed & 1,048 & 0.286 & \textbf{87.08} & 1.27 & 61.75 & \textbf{24.13} & \textbf{73.29} & \textbf{28.91} \\
w/o Task Requirement & 1,048 & \textbf{0.278} & 50.39 & 1.41 & 61.25 & 22.31 & 75.83 & 26.86 \\
w/o Environment State & 1,048 & 0.287 & 82.68 & 1.24 & 61.00 & 19.06 & 77.87 & 24.85 \\
w/o External Information & 1,048 & 0.285 & 80.58 & \textbf{0.31} & \textbf{64.50} & 21.19 & 76.54 & 26.45 \\
\midrule
Mixed & 524 & 0.423 & 80.89 & 8.97 & 59.75 & 22.81 & 74.25 & 27.64 \\
\bottomrule
\end{tabular}
\end{table}
\section{Conclusion}
We presented \name{}, an automated framework for transforming existing static tasks into verifiable dynamic scenarios. The framework generates exogenous events involving task requirements, environment state, and external information, controls their execution, and adapts the original verifier to the changes that actually occur. It preserves the original tools and unaffected task requirements, allowing existing environments to support training under changing conditions. Supervised fine-tuning on 2,095 verified trajectories collected across 36 environments improves the overall performance of Qwen3.5 models of all three evaluated sizes on Gaia2, CostBench, and ToolMaze, supporting generalization beyond the training environments. Our analyses further show diminishing gains from increasing data quantity and highlight the value of including all three event sources for task completion and recovery. However, higher task completion does not always coincide with better execution efficiency, and responding to changes at the appropriate time remains challenging. Future work should investigate how scenario construction and training can better address these limitations. Overall, \name{} provides a practical way to use existing task collections to train agents not only to execute plans, but also to revise their actions as conditions change.

\subsection*{AI use statement}

We used large language models to generate synthetic data, as described in the paper, and to improve the language and clarity of the manuscript. The authors take full responsibility for the content of the paper, including the generated data, experimental results, and conclusions.
\bibliography{tex/iclr2027_conference}

@misc{zhou2024webarenarealisticwebenvironment,
      title={WebArena: A Realistic Web Environment for Building Autonomous Agents}, 
      author={Shuyan Zhou and Frank F. Xu and Hao Zhu and Xuhui Zhou and Robert Lo and Abishek Sridhar and Xianyi Cheng and Tianyue Ou and Yonatan Bisk and Daniel Fried and Uri Alon and Graham Neubig},
      year={2024},
      eprint={2307.13854},
      archivePrefix={arXiv},
      primaryClass={cs.AI},
      url={https://arxiv.org/abs/2307.13854}, 
}

@misc{xie2024osworldbenchmarkingmultimodalagents,
      title={OSWorld: Benchmarking Multimodal Agents for Open-Ended Tasks in Real Computer Environments}, 
      author={Tianbao Xie and Danyang Zhang and Jixuan Chen and Xiaochuan Li and Siheng Zhao and Ruisheng Cao and Toh Jing Hua and Zhoujun Cheng and Dongchan Shin and Fangyu Lei and Yitao Liu and Yiheng Xu and Shuyan Zhou and Silvio Savarese and Caiming Xiong and Victor Zhong and Tao Yu},
      year={2024},
      eprint={2404.07972},
      archivePrefix={arXiv},
      primaryClass={cs.AI},
      url={https://arxiv.org/abs/2404.07972}, 
}

@misc{rawles2025androidworlddynamicbenchmarkingenvironment,
      title={AndroidWorld: A Dynamic Benchmarking Environment for Autonomous Agents}, 
      author={Christopher Rawles and Sarah Clinckemaillie and Yifan Chang and Jonathan Waltz and Gabrielle Lau and Marybeth Fair and Alice Li and William Bishop and Wei Li and Folawiyo Campbell-Ajala and Daniel Toyama and Robert Berry and Divya Tyamagundlu and Timothy Lillicrap and Oriana Riva},
      year={2025},
      eprint={2405.14573},
      archivePrefix={arXiv},
      primaryClass={cs.AI},
      url={https://arxiv.org/abs/2405.14573}, 
}

@misc{yao2024taubenchbenchmarktoolagentuserinteraction,
      title={$\tau$-bench: A Benchmark for Tool-Agent-User Interaction in Real-World Domains}, 
      author={Shunyu Yao and Noah Shinn and Pedram Razavi and Karthik Narasimhan},
      year={2024},
      eprint={2406.12045},
      archivePrefix={arXiv},
      primaryClass={cs.AI},
      url={https://arxiv.org/abs/2406.12045}, 
}

@misc{trivedi2024appworldcontrollableworldapps,
      title={AppWorld: A Controllable World of Apps and People for Benchmarking Interactive Coding Agents}, 
      author={Harsh Trivedi and Tushar Khot and Mareike Hartmann and Ruskin Manku and Vinty Dong and Edward Li and Shashank Gupta and Ashish Sabharwal and Niranjan Balasubramanian},
      year={2024},
      eprint={2407.18901},
      archivePrefix={arXiv},
      primaryClass={cs.SE},
      url={https://arxiv.org/abs/2407.18901}, 
}

@misc{lu2025toolsandboxstatefulconversationalinteractive,
      title={ToolSandbox: A Stateful, Conversational, Interactive Evaluation Benchmark for LLM Tool Use Capabilities}, 
      author={Jiarui Lu and Thomas Holleis and Yizhe Zhang and Bernhard Aumayer and Feng Nan and Felix Bai and Shuang Ma and Shen Ma and Mengyu Li and Guoli Yin and Zirui Wang and Ruoming Pang},
      year={2025},
      eprint={2408.04682},
      archivePrefix={arXiv},
      primaryClass={cs.CL},
      url={https://arxiv.org/abs/2408.04682}, 
}

@misc{barres2025tau2benchevaluatingconversationalagents,
      title={$\tau^2$-Bench: Evaluating Conversational Agents in a Dual-Control Environment}, 
      author={Victor Barres and Honghua Dong and Soham Ray and Xujie Si and Karthik Narasimhan},
      year={2025},
      eprint={2506.07982},
      archivePrefix={arXiv},
      primaryClass={cs.AI},
      url={https://arxiv.org/abs/2506.07982}, 
}

@misc{fang2025generalagenticintelligenceenvironment,
      title={Towards General Agentic Intelligence via Environment Scaling}, 
      author={Runnan Fang and Shihao Cai and Baixuan Li and Jialong Wu and Guangyu Li and Wenbiao Yin and Xinyu Wang and Xiaobin Wang and Liangcai Su and Zhen Zhang and Shibin Wu and Zhengwei Tao and Yong Jiang and Pengjun Xie and Fei Huang and Jingren Zhou},
      year={2025},
      eprint={2509.13311},
      archivePrefix={arXiv},
      primaryClass={cs.CL},
      url={https://arxiv.org/abs/2509.13311}, 
}

@misc{froger2025arescalingagentenvironments,
      title={ARE: Scaling Up Agent Environments and Evaluations}, 
      author={Romain Froger and Pierre Andrews and Matteo Bettini and Amar Budhiraja and Ricardo Silveira Cabral and Virginie Do and Emilien Garreau and Jean-Baptiste Gaya and Hugo Laurençon and Maxime Lecanu and Kunal Malkan and Dheeraj Mekala and Pierre Ménard and Gerard Moreno-Torres Bertran and Ulyana Piterbarg and Mikhail Plekhanov and Mathieu Rita and Andrey Rusakov and Vladislav Vorotilov and Mengjue Wang and Ian Yu and Amine Benhalloum and Grégoire Mialon and Thomas Scialom},
      year={2025},
      eprint={2509.17158},
      archivePrefix={arXiv},
      primaryClass={cs.AI},
      url={https://arxiv.org/abs/2509.17158}, 
}

@misc{song2026envscalerscalingtoolinteractiveenvironments,
      title={EnvScaler: Scaling Tool-Interactive Environments for LLM Agent via Programmatic Synthesis}, 
      author={Xiaoshuai Song and Haofei Chang and Guanting Dong and Yutao Zhu and Ji-Rong Wen and Zhicheng Dou},
      year={2026},
      eprint={2601.05808},
      archivePrefix={arXiv},
      primaryClass={cs.CL},
      url={https://arxiv.org/abs/2601.05808}, 
}

@misc{wang2026agentworldmodelinfinity,
      title={Agent World Model: Infinity Synthetic Environments for Agentic Reinforcement Learning}, 
      author={Zhaoyang Wang and Canwen Xu and Boyi Liu and Yite Wang and Siwei Han and Zhewei Yao and Huaxiu Yao and Yuxiong He},
      year={2026},
      eprint={2602.10090},
      archivePrefix={arXiv},
      primaryClass={cs.AI},
      url={https://arxiv.org/abs/2602.10090}, 
}

@misc{froger2026gaia2benchmarkingllmagents,
      title={Gaia2: Benchmarking LLM Agents on Dynamic and Asynchronous Environments}, 
      author={Romain Froger and Pierre Andrews and Matteo Bettini and Amar Budhiraja and Ricardo Silveira Cabral and Virginie Do and Emilien Garreau and Jean-Baptiste Gaya and Hugo Laurençon and Maxime Lecanu and Kunal Malkan and Dheeraj Mekala and Pierre Ménard and Gerard Moreno-Torres Bertran and Ulyana Piterbarg and Mikhail Plekhanov and Mathieu Rita and Andrey Rusakov and Vladislav Vorotilov and Mengjue Wang and Ian Yu and Amine Benhalloum and Grégoire Mialon and Thomas Scialom},
      year={2026},
      eprint={2602.11964},
      archivePrefix={arXiv},
      primaryClass={cs.AI},
      url={https://arxiv.org/abs/2602.11964}, 
}

@misc{dong2026agentworldscalingrealworldenvironment,
      title={Agent-World: Scaling Real-World Environment Synthesis for Evolving General Agent Intelligence}, 
      author={Guanting Dong and Junting Lu and Junjie Huang and Wanjun Zhong and Longxiang Liu and Shijue Huang and Zhenyu Li and Yang Zhao and Xiaoshuai Song and Xiaoxi Li and Jiajie Jin and Yutao Zhu and Hanbin Wang and Fangyu Lei and Qinyu Luo and Mingyang Chen and Zehui Chen and Jiazhan Feng and Ji-Rong Wen and Zhicheng Dou},
      year={2026},
      eprint={2604.18292},
      archivePrefix={arXiv},
      primaryClass={cs.AI},
      url={https://arxiv.org/abs/2604.18292}, 
}

@misc{meng2026clawmarklivingworldbenchmarkmultiturn,
      title={ClawMark: A Living-World Benchmark for Multi-Turn, Multi-Day, Multimodal Coworker Agents}, 
      author={Fanqing Meng and Lingxiao Du and Zijian Wu and Guanzheng Chen and Xiangyan Liu and Jiaqi Liao and Chonghe Jiang and Zhenglin Wan and Jiawei Gu and Pengfei Zhou and Rui Huang and Ziqi Zhao and Shengyuan Ding and Ailing Yu and Bo Peng and Bowei Xia and Hao Sun and Haotian Liang and Ji Xie and Jiajun Chen and Jiajun Song and Liu Yang and Ming Xu and Qionglin Qiu and Runhao Fu and Shengfang Zhai and Shijian Wang and Tengfei Ma and Tianyi Wu and Weiyang Jin and Yan Wang and Yang Dai and Yao Lai and Youwei Shu and Yue Liu and Yunzhuo Hao and Yuwei Niu and Jinkai Huang and Jiayuan Zhuo and Zhennan Shen and Linyu Wu and Hannah Yao and Charles Chen and Cihang Xie and Yuyin Zhou and Jiaheng Zhang and Zeyu Zheng and Mengkang Hu and Michael Qizhe Shieh},
      year={2026},
      eprint={2604.23781},
      archivePrefix={arXiv},
      primaryClass={cs.CV},
      url={https://arxiv.org/abs/2604.23781}, 
}

@misc{liu2026costbenchevaluatingmultiturncostoptimal,
      title={CostBench: Evaluating Multi-Turn Cost-Optimal Planning and Adaptation in Dynamic Environments for LLM Tool-Use Agents}, 
      author={Jiayu Liu and Cheng Qian and Zhaochen Su and Qing Zong and Shijue Huang and Bingxiang He and Yi R. Fung},
      year={2026},
      eprint={2511.02734},
      archivePrefix={arXiv},
      primaryClass={cs.AI},
      url={https://arxiv.org/abs/2511.02734}, 
}

@misc{zhu2026toolsfailbenchmarkingdynamic,
      title={When Tools Fail: Benchmarking Dynamic Replanning and Anomaly Recovery in LLM Agents}, 
      author={Dongsheng Zhu and Xuchen Ma and Yucheng Shen and Xiang Li and Yukun Zhao and Shuaiqiang Wang and Lingyong Yan and Dawei Yin},
      year={2026},
      eprint={2606.05806},
      archivePrefix={arXiv},
      primaryClass={cs.AI},
      url={https://arxiv.org/abs/2606.05806}, 
}

@misc{debenedetti2024agentdojodynamicenvironmentevaluate,
      title={AgentDojo: A Dynamic Environment to Evaluate Prompt Injection Attacks and Defenses for LLM Agents}, 
      author={Edoardo Debenedetti and Jie Zhang and Mislav Balunović and Luca Beurer-Kellner and Marc Fischer and Florian Tramèr},
      year={2024},
      eprint={2406.13352},
      archivePrefix={arXiv},
      primaryClass={cs.CR},
      url={https://arxiv.org/abs/2406.13352}, 
}

@misc{cai2025autoforgeautomatedenvironmentsynthesis,
      title={AutoForge: Automated Environment Synthesis for Agentic Reinforcement Learning}, 
      author={Shihao Cai and Runnan Fang and Jialong Wu and Baixuan Li and Xinyu Wang and Yong Jiang and Liangcai Su and Liwen Zhang and Wenbiao Yin and Zhen Zhang and Fuli Feng and Pengjun Xie and Xiaobin Wang},
      year={2025},
      eprint={2512.22857},
      archivePrefix={arXiv},
      primaryClass={cs.CL},
      url={https://arxiv.org/abs/2512.22857}, 
}

@misc{ji2026clawarenabenchmarkingaiagents,
      title={ClawArena: Benchmarking AI Agents in Evolving Information Environments}, 
      author={Haonian Ji and Kaiwen Xiong and Siwei Han and Peng Xia and Shi Qiu and Yiyang Zhou and Jiaqi Liu and Jinlong Li and Bingzhou Li and Zeyu Zheng and Cihang Xie and Huaxiu Yao},
      year={2026},
      eprint={2604.04202},
      archivePrefix={arXiv},
      primaryClass={cs.LG},
      url={https://arxiv.org/abs/2604.04202}, 
}

@misc{seo2026simuhometemporalenvironmentawarebenchmark,
      title={SimuHome: A Temporal- and Environment-Aware Benchmark for Smart Home LLM Agents}, 
      author={Gyuhyeon Seo and Jungwoo Yang and Junseong Pyo and Nalim Kim and Jonggeun Lee and Yohan Jo},
      year={2026},
      eprint={2509.24282},
      archivePrefix={arXiv},
      primaryClass={cs.CL},
      url={https://arxiv.org/abs/2509.24282}, 
}

@misc{li2026worldwontstaystill,
      title={The World Won't Stay Still: Programmable Evolution for Agent Benchmarks}, 
      author={Guangrui Li and Yaochen Xie and Yi Liu and Ziwei Dong and Xingyuan Pan and Tianqi Zheng and Jason Choi and Michael J. Morais and Binit Jha and Shaunak Mishra and Bingrou Zhou and Chen Luo and Monica Xiao Cheng and Dawn Song},
      year={2026},
      eprint={2603.05910},
      archivePrefix={arXiv},
      primaryClass={cs.AI},
      url={https://arxiv.org/abs/2603.05910}, 
}

@misc{xu2026evoarenatrackingmemoryevolution,
      title={EvoArena: Tracking Memory Evolution for Robust LLM Agents in Dynamic Environments}, 
      author={Jundong Xu and Qingchuan Li and Jiaying Wu and Yihuai Lan and Shuyue Stella Li and Huichi Zhou and Bowen Jiang and Lei Wang and Jun Wang and Anh Tuan Luu and Caiming Xiong and Hae Won Park and Bryan Hooi and Zhiyuan Hu},
      year={2026},
      eprint={2606.13681},
      archivePrefix={arXiv},
      primaryClass={cs.CL},
      url={https://arxiv.org/abs/2606.13681}, 
}
\bibliographystyle{iclr2027_conference}
\appendix
\section{Detailed Evaluation Results}
\label{app:toolmaze_detailed_results}
Tables~\ref{tab:toolmaze_detail} provide the detailed evaluation results on ToolMaze. For ToolMaze, NP denotes execution without injected failures; P1 and P2 introduce explicit transient and permanent failures, respectively, whereas P3 and P4 introduce implicit transient and permanent output corruption.

\begin{table}[t]
\centering
\setlength{\belowcaptionskip}{\baselineskip}
\scriptsize
\setlength{\tabcolsep}{2.5pt}
\renewcommand{\arraystretch}{1.12}
\caption{
Detailed results on ToolMaze.
}
\label{tab:toolmaze_detail}
\resizebox{\linewidth}{!}{%
\begin{tabular}{l*{17}{r}}
\toprule
 & \multicolumn{6}{c}{TSR $\uparrow$}
 & \multicolumn{5}{c}{PRR $\uparrow$}
 & \multicolumn{5}{c}{RC $\downarrow$}
 & Overall \\
\cmidrule(lr){2-7}
\cmidrule(lr){8-12}
\cmidrule(lr){13-17}
\cmidrule(lr){18-18}
Model
 & NP & P1 & P2 & P3 & P4 & Avg
 & P1 & P2 & P3 & P4 & Avg
 & P1 & P2 & P3 & P4 & Avg
 & Avg $\uparrow$ \\
\midrule

GPT-5.5
 & 70.00 & 67.25 & 26.75 & 16.75 & 14.00 & 31.19
 & 86.09 & 29.49 & 18.89 & 14.75 & 37.31
 & 35.61 & 73.02 & 81.33 & 83.83 & 68.45
 & 35.94 \\

Deepseek-V4-Pro
 & 69.75 & 68.00 & 34.75 & 12.00 & 9.75 & 31.12
 & 84.70 & 37.74 & 12.41 & 12.00 & 36.71
 & 29.98 & 63.55 & 85.19 & 87.55 & 66.57
 & 36.33 \\

GLM-5.1
 & 75.75 & 60.00 & 32.50 & 8.00 & 8.50 & 27.25
 & 67.19 & 35.33 & 9.04 & 10.15 & 30.43
 & 37.86 & 66.47 & 88.65 & 89.11 & 70.52
 & 32.28 \\

Gemini-3.1-Pro-Preview
 & 71.25 & 59.75 & 30.00 & 6.75 & 9.00 & 26.38
 & 73.49 & 33.52 & 8.09 & 10.84 & 31.48
 & 37.53 & 67.64 & 88.64 & 87.24 & 70.26
 & 32.19 \\

GPT-4o
 & 59.25 & 47.00 & 33.50 & 9.25 & 11.50 & 25.31
 & 70.49 & 27.72 & 12.72 & 12.71 & 30.91
 & 35.97 & 78.34 & 81.77 & 80.54 & 69.16
 & 31.29 \\

\midrule

Qwen3.5-2B
 & 42.50 & 24.50 & 14.00 & 5.75 & 4.25 & 12.13
 & 46.70 & 12.10 & 12.08 & 5.00 & 18.97
 & 67.76 & 81.27 & 88.00 & 89.37 & 81.60
 & 18.52 \\

\rowcolor{blue!6}
\quad +\name
 & 53.25 & 49.75 & 22.50 & 7.00 & 9.50 & 22.19
 & 48.81 & 18.37 & 12.85 & 6.46 & 21.62
 & 63.66 & 79.27 & 86.86 & 88.46 & 79.56
 & 23.49 \\

\addlinespace[2pt]

Qwen3.5-4B
 & 57.25 & 45.50 & 23.75 & 9.00 & 6.75 & 21.25
 & 64.68 & 22.49 & 12.13 & 6.74 & 26.51
 & 53.08 & 74.28 & 87.71 & 89.60 & 76.17
 & 26.26 \\

\rowcolor{blue!6}
\quad +\name
 & 65.00 & 54.25 & 27.25 & 8.50 & 11.50 & 25.38
 & 67.49 & 26.34 & 10.22 & 11.52 & 28.89
 & 43.84 & 71.57 & 87.20 & 86.24 & 72.21
 & 29.99 \\

\addlinespace[2pt]

Qwen3.5-9B
 & 64.75 & 56.50 & 31.50 & 7.50 & 10.00 & 26.38
 & 69.06 & 30.13 & 8.76 & 11.39 & 29.83
 & 41.49 & 68.74 & 90.02 & 88.20 & 72.11
 & 30.59 \\

\rowcolor{blue!6}
\quad +\name
 & 66.25 & 56.50 & 31.75 & 14.75 & 10.25 & 28.31
 & 67.91 & 29.75 & 18.36 & 10.21 & 31.56
 & 43.56 & 67.63 & 83.96 & 87.35 & 70.62
 & 32.28 \\

\bottomrule
\end{tabular}%
}
\end{table}

\end{document}